\documentclass{article}
\usepackage[final]{neurips_2022}
\usepackage[utf8]{inputenc} 
\usepackage[T1]{fontenc}    
\usepackage{hyperref}       
\usepackage{url}            
\usepackage{booktabs}       
\usepackage{amsfonts}       
\usepackage{nicefrac}       
\usepackage{microtype}      
\usepackage{amsmath,amssymb,amsfonts}
\usepackage[dvipsnames]{xcolor}
\usepackage{subcaption}
\usepackage{multirow}
\usepackage{fancyhdr}
\usepackage{array}
\usepackage{mathtools}
\usepackage{rotating}
\newcommand\Tstrut{\rule{0pt}{2.35ex}}         
\newcommand\Bstrut{\rule[-0.9ex]{0pt}{0pt}}   
\begin{document}

\title{Beyond CAGE: Investigating Generalization of Learned Autonomous Network Defense Policies}


\author{%
  Melody Wolk, Andy Applebaum, Camron Dennler, Patrick Dwyer, Marina Moskowitz,\\
  \textbf{Harold Nguyen, Nicole Nichols, Nicole Park, Paul Rachwalski,}\\
  \textbf{Frank Rau, and Adrian Webster}\\
  Apple\\
  Cupertino, CA\\ \\%
  \begin{tabular}{lcll}
  \multirow{3}{*}{\{}& \texttt{melody\_wolk, aapplebaum, cdennler, pnd, mmoskowitz,} & \multirow{3}{*}{\}}& \multirow{3}{*}{\texttt{@apple.com}}\\
  & \texttt{harold\_nguyen, nicole\_nichols, n\_park, prachwalski,} & & \\
  & \texttt{frau, adwebster} & & \\
  \end{tabular}
}


\maketitle

\begin{abstract}
Advancements in reinforcement learning (RL) have inspired new directions in intelligent automation of network defense.
However, many of these advancements have either outpaced their application to network security or have not considered the challenges associated with implementing them in the real-world.
To understand these problems, this work evaluates several RL approaches implemented in the second edition of the CAGE Challenge, a public competition to build an autonomous network defender agent in a high-fidelity network simulator.
Our approaches all build on the Proximal Policy Optimization (PPO) family of algorithms, and include hierarchical RL, action masking, custom training, and ensemble RL.
We find that the ensemble RL technique performs strongest, outperforming our other models and taking second place in the competition.
To understand applicability to real environments we evaluate each method's ability to generalize to unseen networks and against an unknown attack strategy.
In unseen environments, all of our approaches perform worse, with degradation varied based on the type of environmental change.
Against an unknown attacker strategy, we found that our models had reduced overall performance even though the new strategy was less efficient than the ones our models trained on.
Together, these results highlight promising research directions for autonomous network defense in the real world.

\end{abstract}


\section{Introduction}

Modern network defense is dominated by human processes, such as alert triage and incident response.
\emph{Playbook automation} \cite{stevens2022ready,iacdplaybook, oasiscacao} can alleviate human cognitive fatigue by standardizing  human-crafted decision logic, but is brittle compared to the fully-automated potential of RL-based network defense agents, as shown in recent RL successes such as Starcraft \cite{vinyals2019grandmaster}, DotA 2 \cite{berner2019dota}, and Stratego \cite{strateg}.
Using RL agents to perform \textbf{fully autonomous} network defense \cite{foley2022autonomous, molina2021network, hammar2020finding, hammar2022learning, ridley2018machine} could allow analysts to instead focus their attention on sophisticated scenarios and improved response in general. 

This work complements current research by analyzing multiple RL approaches -- specifically, variants of Proximal Policy Optimization (PPO) -- as applied to the \emph{CAGE Challenge} \cite{cage_challenge_2_announcement}, a public competition to build automated network defender agents, as well as to real-world integration scenarios.
The challenge uses a high-fidelity network simulator (\emph{CybORG} \cite{cyborg_acd_2021}) that provides a realistic simulation of an attacker on an enterprise network.
Using CybORG, we make the following contributions:

\begin{itemize}
    \item We describe autonomous network defense methods and performance as tested in the CAGE Challenge, including hierarchical RL, ensemble RL, action masking, and transfer learning.
    \item We analyze each of our approaches' ability to generalize to unseen network environments and against an unseen but inefficient attacker strategy.
\end{itemize}
 
Our results demonstrate that ensemble RL is an effective and promising methodology for building an autonomous network defender.
However, they also show that all of our approaches struggle when tested in an unseen environment -- being sensitive to host information and attack path -- or against an unseen attacker, with results aligning with the stochasticity of the attacker.
These contributions both show positive results for new RL methodologies applied to autonomous network defense, as well as identify challenges for real-world integration of such methodologies.


The rest of this paper is organized as follows: Section~\ref{sec:background} describes the CAGE Challenge, CybORG, and related work; Section~\ref{sec:approaches} details our approaches' methodology; Section~\ref{sec:results} provides initial results and analysis; Section~\ref{sec:integration} tests our approaches in scenarios that emulate real-world challenges; Section~\ref{sec:future} outlines areas for future work; and Section~\ref{sec:discussion} summarizes main points.
Additionally, Appendix~\ref{app:params} includes details on model hyper-parameters, training schedules, and additional metrics.


\section{Background}\label{sec:background}
The problem of autonomous network defense is to determine the optimal response given a series of observations -- typically \emph{intrusion alerts} -- from an enterprise network.
These observations contain both false positives -- i.e., benign user activity erroneously flagged as malicious -- and false negatives, i.e. the alerting components not identifying the attacker, of which the defender needs to balance.

Research to solve this problem includes applying heuristics \cite{carver2000methodology}, expert systems \cite{porras1997emerald}, model-based reasoning \cite{toth2002evaluating}, game theory \cite{zonouz2013rre}, and formal planning \cite{musman2019steps}.
Recent research has also suggested to use RL; early work \cite{ridley2018machine, elderman2017adversarial} used highly abstracted network simulators to experiment with multiple RL approaches, including Tabular Q-Learning, Upper Confidence Bound 1 \cite{auer2002finite}, Discounted Upper Confidence Bound \cite{garivier2008upper}, and others.
Other approaches moved past pure network simulation towards emulation: \cite{molina2021network} used containers to emulate a real network, testing RL techniques Ape-X Deep Q-Networks (APEX-DQN) \cite{horgan2018distributed}, PPO \cite{schulman2017proximal}, and Asynchronous Advantage Actor-Critic (A3C) \cite{mnih2016asynchronous}.
Most of these works used RL techniques without modification, though \cite{hammar2020finding} tests PPO, REINFORCE \cite{reinforce}, and a custom auto-regressive PPO, all within an abstract network simulator.
In almost all cases, the RL approaches achieved success, bringing the added benefit that these approaches are able to \emph{learn} the dynamics of the defended network without an explicit blueprint.

\subsection{CybORG}\label{subsec:cyborg}

\begin{figure*}
    \centering
    \includegraphics[width=.8\textwidth]{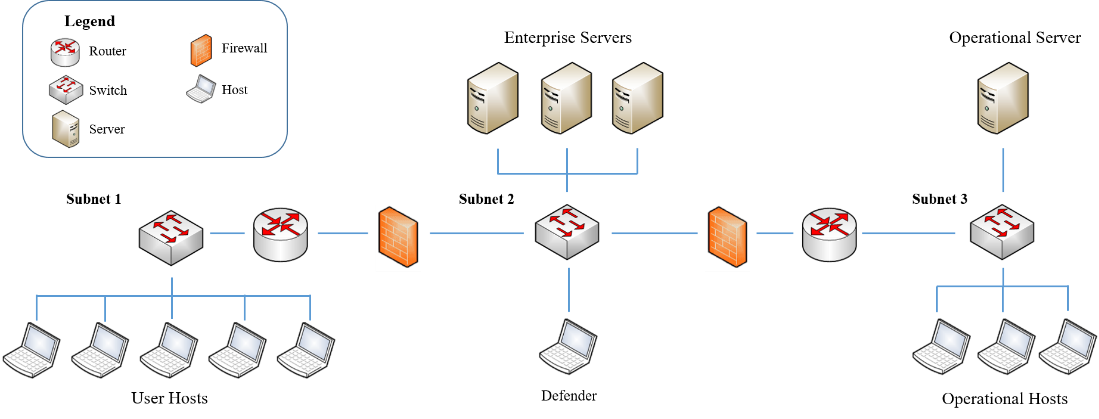}
    \caption{Visualization of CybORG Scenario 2 (taken from \cite{cage_challenge_2_announcement}). Subnet 1 consists of user hosts; Subnet 2 of important enterprise servers; and Subnet 3 of a critical operational server and three user hosts.}
    \label{fig:scenario2}
\end{figure*}

The CybORG simulator provides an interface for attacker and defender agents to interact with a complex network simulation environment, allowing for detailed encoding of real-world scenarios that is abstracted through a series of wrappers enabling an RL interface while maintaining good realism.
Specific environments are encoded as \emph{scenarios}, with the second edition of the CAGE Challenge using \emph{Scenario 2} (Figure~\ref{fig:scenario2}).
Simulations in CybORG are played over these scenarios as \emph{games} between the attacker and defender agent, and also feature a \emph{green agent} that simulates user activity and generates false positives for the defender.
Each turn the attacker agent can scan hosts and subnets, launch exploits for lateral movement, escalate privileges, or disrupt a compromised host.
In response, the defender can do nothing (\emph{Sleep}); \emph{Monitor} the entire network for malicious activity; \emph{Analyze} a specific host to identify malicious artifacts; \emph{Remove} known malicious artifacts from a host; \emph{Restore} a host to a known good state and remove any attacker foothold; or instantiate one of seven \emph{Decoy} services on the target host.
Each turn the defender receives an observation of the current network state as a \emph{bit-vector} -- where each host is represented as a fixed set of bits -- and a numeric reward.
The value of the reward is determined by the severity of the attacker's access to each host, with the defender also receiving a penalty each time it uses the \emph{Restore} action; the reward is always negative such that a game score of 0 indicates a perfect defense.

The CybORG environment allows construction of attacker agents; e.g. \cite{tran2021deep} used CybORG to show how hierarchical deep RL can build a better attacker agent than DQN.
For CAGE, the attacker is restricted to using one of three built-in fixed-approach strategies.
The first is the \emph{BLine} attacker agent, which is pre-configured with full knowledge of the network .
The second is the \emph{Meander} attacker agent, which needs to discover network details during the game, and seeks to compromise each subnet before moving to the next one.
The last is a \emph{Sleep} agent, which executes a no-op each step.

\subsection{CAGE}\label{subsec:cage}
In the CAGE Challenge \cite{cage_challenge_2_announcement}, participants submit defender agents that are ran within the CybORG simulator.
For evaluation, each agent is run against the three attackers mentioned above at three different game lengths (30, 50, and 100).
The agent's final score is the summation of all rewards received against each attacker at each step length.
This construction gives CAGE two primary challenges: (1) identify an optimal policy given noisy network observations, and (2) defend the network without explicitly knowing which attacker strategy is in use.
To accompany the challenge, baseline results are provided for a PPO-based agent, a Heuristic agent using only the Restore action, an agent that chooses actions randomly, and a ``Sleep'' agent. Table~\ref{tab:baselines} reproduces these results, with the PPO agent significantly outperforming the other baselines by all measures.

\begin{table*}\footnotesize
    \centering
    \begin{tabular}{|l|c|cc|cc|cc|}
        \hline
        \multirow{2}{*}{Method} & \multirow{2}{*}{\shortstack[c]{Total\\Reward}} & \multicolumn{2}{c|}{30 Steps} & \multicolumn{2}{c|}{50 Steps} & \multicolumn{2}{c|}{100 Steps}\Tstrut\\
        & & BLine & Meander & BLine & Meander & BLine & Meander\Bstrut\\
        \hline
        PPO Baseline & -95.47 & -8.75 & -7.31 & -14.92 & -12.04 & -30.19 & -22.21 \Tstrut\\
        Heuristic Restore & -823.89& -58.33 & -44.29 & -93.05 & -85.75& -184.34 & -192.63\\
        Random & -2015.61& -154.06 & -33.43 & -347.1 & -171.49 & -726.92 & -566.41 \\
        Sleep & -3112.2& -218.65 & -39.31& -480.17 & -267.6 & -1134.03 & -972.43\Bstrut\\
        \hline
    \end{tabular}
    \caption{Baseline results provided by the CAGE Competition organizers \cite{cage_challenge_2_announcement}. The defender seeks to maximize the total reward, with a best possible total of 0, indicating the attacker has established no footholds and the defender has not taken any Restore actions. Note that we omit the \emph{Sleep} attacker in the interest of space.}
    \label{tab:baselines}
\end{table*}

\cite{foley2022autonomous} describes the solution to the winner of the first edition of the CAGE Challenge \cite{cage_challenge_1} using Hierarchical PPO (HiPPO).
They train two PPO policies, one against BLine and another against Meander.
They then train a third policy that seeks only to deploy the pre-trained BLine or Meander policies.

\section{Approaches}\label{sec:approaches}
Each of our approaches build on Proximal Policy Optimization (PPO) \cite{schulman2017proximal} as the core RL algorithm.
All models were implemented using the RLlib \cite{liang2018rllib} or Stable Baselines3 \cite{stable-baselines3} libraries, which both provide access to PPO out-of-the-box; unless otherwise mentioned, we use RLlib by default.
For exploration, models built with RLlib use Curiosity \cite{pathak2017curiosity} and for Stable Baselines3 use the default Gaussian noise mechanism.
Tables \ref{tab:trainingschedules} and \ref{tab:allhyperparameters} in Appendix~\ref{app:params} provide additional model information.

\paragraph{Attacker Randomization}

To avoid the defender from over-fitting to a specific attacker strategy we varied the specific attacker during training.
Each time an episode ends and the environment is reset, with probability $Pr(\texttt{BLine})$ we initialize the environment with the BLine attacker, and with probability $Pr(\texttt{Meander}) = 1 - Pr(\texttt{BLine})$ we initialize the environment with the Meander attacker.
This idea factored into many of our approaches discussed later in this section.
Additionally, we also trained a single PPO model with Stable Baselines3 and Optuna for hyper-parameter search against an environment with $Pr(\texttt{BLine}) = Pr(\texttt{Meander}) = 0.50$; we refer to this model as \textbf{\emph{Tuned PPO}}.


\paragraph{Transfer Learning}
Another approach to combat attacker strategy uncertainty is to use a simple transfer learning procedure where we continually train the same agent on different CybORG configurations, each in sequence.
These configurations vary which attacker agent the defender is trained against as well as how many steps each game lasts.
Additionally, each iteration features a grid search over several PPO and Curiosity hyper-parameters, with the best performing configuration used to continue training and the others discarded.
We refer to this model as \textbf{\emph{Transfer Learning}}.

\paragraph{Action Masking}
Action masking is a common technique used to avoid executing known invalid actions.
Prior work \cite{DBLP:journals/corr/abs-2006-14171} has shown that even when the mask is removed, the agent is still able to produce useful behaviors. 
We explored multiple approaches to leverage this idea, ultimately settling on one that used all actions but learned an embedding during training.
Additionally, we used a similar training schedule as for Transfer Learning alongside an attacker randomization procedure to avoid overfitting each iteration.
We refer to the resulting approach as \textbf{\emph{TL + Embedding}}.

\paragraph{Hierarchical PPO}
The idea behind this approach is similar to \cite{hammar2020finding}, where an action is decided, followed by a host to perform that action on.
Our implementation differs as we use the RLlib multi-agent training environment as opposed to an auto-regressive model: in our approach, we implement a \emph{high-level} policy that identifies what \emph{action} to take, and a \emph{low-level} policy that identifies what \emph{target} to take that action against.
To ensure attacker diversity during training, each time an episode ends it is reset with the BLine, Meander, or Sleep attackers, picking each in sequence.
We refer to this model as \textbf{\emph{HPPO}}, differing from the other hierarchical \emph{HiPPO} method \cite{foley2022autonomous} mentioned earlier.


\paragraph{Ensemble RL}

Our last approach uses a simple ensemble technique consisting of individual PPO models trained with different attacker randomization settings and various PPO hyper-parameters.
Our ensemble uses weighted majority voting \cite{wiering2008ensemble}, choosing the action preferred by most models and breaking ties by weighing votes by each model's score; these scores are generated beforehand by testing each model against the CAGE Challenge procedure, assigning the result as the model's score.
Using this approach we generate two agents for the CAGE Challenge.
The first, \textbf{\emph{Ensemble}}, is one of the higher-performing ensembles that we constructed and consists of seven different PPO models.
The second, \textbf{\emph{Multi-Ensemble}}, uses the same idea but consists of five distinct \emph{ensembles} as opposed to individual models.
Appendix~\ref{app:params}, Table~\ref{tab:allensembles} provides more information about each ensemble.

\section{Results}\label{sec:results}

\begin{table*}\renewcommand{\arraystretch}{1}\setlength{\tabcolsep}{5.95pt}\footnotesize
    \centering
    \begin{tabular}{|l|c|cc|cc|cc||c|}
        \hline
        \multirow{2}{*}{Method} &\multirow{2}{*}{\shortstack[c]{Total\\Reward}} & \multicolumn{2}{c|}{30 Steps} & \multicolumn{2}{c|}{50 Steps} & \multicolumn{2}{c||}{100 Steps} & \multirow{2}{*}{\shortstack[c]{CAGE\\Place}}\Tstrut\\
         & & BLine & Meander & BLine & Meander & BLine & Meander & \Bstrut\\
        \hline
        Multi-Ensemble & -57.14 & -3.44 & -6.46 & -6.07 & -10.22 & -11.5 & -19.45 &2\Tstrut\\
        Ensemble & -58.62 & -3.6 & -6.55 & -6.14 & -10.45 & -12.08 & -19.81 & 6\\
        Transfer Learning & -69.91 & -5.6 & -6.43 & -9.5 & -10.34 & -19.82 & -18.24& 8\\
        TL + Embedding& -72.26  & -4.49 & -6.31 & -8.07 & -11.36 & -17.72 & -24.31& 9\\
        Tuned PPO & -73.44 & -4.74 & -8.06 & -8.17 & -12.6 & -17.04 & -22.82 & 10\\
        HPPO & -120.15 & -4.68 & -9.16 & -8.45 & -22.25 & -17.72 & -57.89 & 14\Bstrut\\
        \hline
        Full Visibility & -54.73 & -3.9 & -5.47 & -6.97 & -8.27 & -14.94 & -15.18 & - \Tstrut\Bstrut\\
        \hline
    \end{tabular}
    \caption{Results for methods in Section~\ref{sec:approaches} after running 1000 episodes against BLine and Meander and with game lengths of 30, 50, and 100; \emph{Total Reward} shows the aggregated result across each step number and attacker agent. Additional information on variance and seed sensitivity can be found in Appendix~\ref{app:variance}. The right-most column shows the placement of each model within the CAGE Challenge, with full results available at \cite{cage_challenge_2_announcement}.}
    \label{tab:finalresults}
\end{table*}

We test each of our approaches using a similar process as in the CAGE Challenge itself, aggregating the average total reward over 1000 episodes against BLine and Meander at game lengths of 30, 50, and 100. Data from game lengths 30 and 50 are unique and not derived from intermediate results of 100. 
Table~\ref{tab:finalresults} presents our results, with \emph{Total Reward} showing the result for each defensive approach.
Overall, the Ensemble and Multi-Ensemble approaches score best at -58.62 and -57.14 respectively.
Given the best possible reward is 0, Multi-Ensemble appears strongest; while we found a high variance based on seed for these results, anecdotally Multi-Ensemble almost uniformly edges out Ensemble by a point or two.
Both approaches perform quite well against BLine, scoring five points better than any non-ensemble approach in the 100 Steps case.
Against Meander, we find both of their scores within a few points of our other approaches, in some cases even being behind another.

On the other end of the spectrum, HPPO performs relatively poorly, scoring worse than the provided PPO baseline in Table~\ref{tab:baselines} and the least of our approaches.
Interestingly, while this approach performed poor in aggregate, its score against BLine was on par with TL + Embedding and Tuned PPO, and outperformed Transfer Learning.
With additional tuning, such as a more exhaustive hyper-parameter search or using an alternative training schedule, this approach might have been stronger.

Transfer Learning, TL + Embedding, and Tuned PPO all performed well, scoring -69.91, -72.26, and -73.44 respectively.
Of the three, Transfer Learning had the most notable results, outperforming both Ensemble and Multi-Ensemble for Meander at 30 and 100 Steps.
This strong performance is likely due to the high ratio of trials trained against Meander versus BLine; while Transfer Learning was the strongest approach against Meander, it was one of the worst against BLine.
TL + Embedding, despite scoring close to Transfer Learning in aggregate, had a very different profile, performing well against BLine but struggling against Meander.
Interestingly, even though it scored best among our approaches against Meander at 30 Steps, this model was one of the worst against Meander at 100 Steps.
This difference between the two may be due to the training schedule: Transfer Learning trained exclusively on BLine \emph{or} Meander, whereas TL + Embedding used a percentage-wise split.


In the remainder of this section we attempt to better elicit the efficacy of each of our approaches, comparing their performance in the CAGE Challenge as well as to a defender that has full visibility of the attacker's state.
We also looked at the action distribution of each approach; this did not reveal any particular insights, but we include it in Appendix ~\ref{app:actdist} for interested readers.
We do note that based on this analysis future work should seek to examine \emph{explainability} for RL-based approaches.

\paragraph{Performance in CAGE}
Table~\ref{tab:finalresults} in the right-most column shows the placement for each of our models within the CAGE Challenge.
Multi-Ensemble did best, taking second place to team \emph{CardiffUniv}'s HiPPO implementation, which scored -54.57 for clear first.
Third, fourth, and fifth place were from the authors of \cite{foley2022autonomous}, each using a HiPPO variation and scoring -57.05, \mbox{-57.29}, and \mbox{-57.46}.
The seventh place team (\emph{UoA}) had a score of -69.88 and used a Belief Ensemble approach.
Based on these results and the competitiveness of the Challenge, we felt that our models performed well.

\paragraph{Full Visibility}
As another point of comparison, we train a model under conditions with no noise -- i.e., no false positives or negatives -- and knowledge of the attacker strategy, using PPO-with-Curiosity in RLlib with mostly default hyper-parameters and assuming a top-level model that deploys the correct sub-model.
We refer to this model as \textbf{\emph{Full Visibility}}, with results shown in the last row of Table~\ref{tab:finalresults}.
Compared to this new model, our approaches perform well: Ensemble and Multi-Ensemble are both within a few points of Full Visibility and the others are not far behind; in fact, both ensemble approaches \emph{outperform} Full Visibility against the BLine attacker.
This is likely explained by the relative determinism of each attacker strategy: as BLine follows a fixed path, it is easier to anticipate its movements, and therefore the impact of no false positives or 
negatives is less significant.
That our models perform so closely to this agent suggests that they do offer good performance.


\section{Real World Integration}\label{sec:integration}
Our results strongly suggest that various RL techniques can be used to construct an effective autonomous network defense agent.
However, in practice integrating an RL-based defender into a real system is fraught with complexities. 
For one, training an agent and running live attacks against a production network can significantly impact operations.
An alternative is to train on an \emph{emulated} network setup to mimic the production network, but this is not without drawbacks either: \cite{li2021cygil} trained an automated attacker agent in a fully virtualized network with only a small number of hosts, finding the length of a full episode ranging from 20 to 30 minutes.
At one million trials, the required time to train would take on the order of years; \emph{distributed training} can reduce this, but brings with it different costs.
An effective alternative is to train the defender within a high-fidelity simulator, as simulations (1) do not impact the production network; (2) can run very quickly; and (3) do not have stringent hardware requirements.
However, this approach requires that the simulator maintains an accurate representation of the target environment for the RL agent to train in.
In this section, we explore integration challenges of a simulation-based approach to (1) make timely decisions; (2) generalize to unseen network environments; and (3) provide robustness against unseen attacker strategies.


\subsection{Computation Time}\label{subsec:timing}
\begin{table}\renewcommand{\arraystretch}{1}\setlength{\tabcolsep}{6pt}\footnotesize
    \centering
    \begin{tabular}{|l|c|p{2.5cm}p{2.5cm}|}
        \hline
        \multirow{2}{*}{Model} & \multirow{2}{*}{\shortstack[c]{Total Average\\Decision Time (ms)}} & \multicolumn{2}{p{5.35cm}|}{\centering Avg. Decision Time by Attacker (ms)} \Tstrut\\
        & & \multicolumn{1}{p{2.5cm}}{\centering BLine}  & \multicolumn{1}{p{2.5cm}|}{\centering Meander} \Bstrut\\
        \hline
        Heuristic Restore & 0.006 $\pm$ 0.001 & \multicolumn{1}{p{2.5cm}}{\centering 0.006} & \multicolumn{1}{p{2.5cm}|}{\centering 0.006}\Tstrut\Bstrut\\
        \hline
        Multi-Ensemble & 27.80 $\pm$ 3.9 & \multicolumn{1}{p{2.5cm}}{\centering 28.82} & \multicolumn{1}{p{2.5cm}|}{\centering 27.42}\Tstrut\\
        Ensemble & 6.67 $\pm$ 0.38 & \multicolumn{1}{p{2.5cm}}{\centering 6.67} & \multicolumn{1}{p{2.5cm}|}{\centering 6.68}\\
        Transfer Learning & 0.81 $\pm$ 0.07 & \multicolumn{1}{p{2.5cm}}{\centering 0.81} & \multicolumn{1}{p{2.5cm}|}{\centering 0.80}\\
        TL + Embedding & 0.70 $\pm$ 0.1 & \multicolumn{1}{p{2.5cm}}{\centering 0.71} & \multicolumn{1}{p{2.5cm}|}{\centering 0.69}\\
        Tuned PPO & 0.27 $\pm$ 0.05 & \multicolumn{1}{p{2.5cm}}{\centering 0.27} & \multicolumn{1}{p{2.5cm}|}{\centering 0.26}\\
        HPPO & 1.12 $\pm$ 0.47 & \multicolumn{1}{p{2.5cm}}{\centering 1.38} & \multicolumn{1}{p{2.5cm}|}{\centering 1.01}\Bstrut\\
        \hline
    \end{tabular}
    \caption{\Tstrut Average time to compute a single action in milliseconds, with the total listed with a 95\% confidence interval. Averaged over 10000 decisions, consisting of 500 games of length 100 against both BLine and Meander.}
    \label{tab:timing}
\end{table}
Table~\ref{tab:timing} lists the average time it took in milliseconds for each model to make a decision, computed over 500 episodes of length 100 against BLine and Meander.
The Heuristic Restore baseline is clearly the fastest, taking only 6 \emph{micro}seconds, well below the RL models.
This is expected as this approach only looks at the previous observation, and is akin to e.g. responding immediately to a known alert.
Within the RL approaches, Tuned PPO is clearly fastest at 0.27ms.
TL + Embedding and Transfer Learning both take a similar amount of time at 0.7ms and 0.81ms respectively.
We suspect Tuned PPO's speed advantage is due to Stable Baselines3 evaluating more quickly than RLlib.
The last three approaches all take longer, with HPPO quickest at 1.12ms, Multi-Ensemble slowest at 27.8ms, and Ensemble between the two at 6.67ms.
This distribution aligns with the number of constituent models each of these three has, and is consistent with multiplying the average time for e.g. Transfer Learning with the number of models within each approach (two for HPPO, seven for Ensemble, and 37 for Multi-Ensemble).
Looking at the rightmost two columns of the table, we do not see any significant difference in decision time based on attacker for any of the models.

As all of the RL approaches average well under a second, it does not appear that computation time would be a significant challenge for deployment.
However, with some changes -- such as a larger network or action/observation space, or if Multi-Ensemble grows to use thousands of models -- timing may be more impactful; in these cases, distributed evaluation can help speed up the computation.

\subsection{Topology Generalization}\label{subsec:topology}
\begin{table*}\footnotesize\setlength{\tabcolsep}{4.5pt}
    \centering
    \begin{tabular}{|l|l|c|c|cc|cc|cc|}
    \hline
    \multirow{2}{*}{} & \multirow{2}{*}{Model} & \multirow{2}{*}{\shortstack[c]{Total\\Reward}} & \multirow{2}{*}{\shortstack[c]{Percent\\Change}} & \multicolumn{2}{c|}{30 Steps} & \multicolumn{2}{c|}{50 Steps} & \multicolumn{2}{c|}{100 Steps}\Tstrut\\
    && & & BLine & Meander & BLine & Meander & BLine & Meander\Bstrut\\
    \hline
    \multirow{6}{*}{\rotatebox[origin=c]{90}{\shortstack[c]{Scenario 3:\\Image + Path}}} & Multi-Ensemble & -166.2 & \textcolor{RawSienna}{\textbf{-190.8}} & -13 &-8.5 &-22.9 &-14.4 &-54.5 &-27.6\Tstrut\\
    & Ensemble & -147.2 & \textcolor{RawSienna}{\textbf{-154.5}} & -13.1 & -8.7 & -16 & -14.5& -56.9& -28\\
    & Transfer Learning & -131.9 & \textcolor{RawSienna}{\textbf{-88.7}} & -13.5&-8.1 &-23.5 & -13.4& -50.6&-22.8\\
    & TL + Embedding & -166.2 & \textcolor{RawSienna}{\textbf{-129.9}} & -14.9 &-8.5 &-27.9 &-16.5 &-61.5 &-36.8\\
    & Tuned PPO & -95.2 & \textcolor{RawSienna}{\textbf{-29.6}} & -8.7 & -8.3 & -14.5 & -13 & -27.3 & -23.4\\
    & HPPO & -176.9 & \textcolor{RawSienna}{\textbf{-47.3}} & -11.8 & -9.1 &-23.2 &-22.1 &-55.2 & -55.6\Bstrut\\
    \hline
    \multirow{6}{*}{\rotatebox[origin=c]{90}{\shortstack[c]{Scenario 4:\\Image}}} & Multi-Ensemble & -172.3 & \textcolor{RawSienna}{\textbf{-201.6}} & -14.4 & -10 & -26.3 & -20.2 & -56.3 & -45.1\Tstrut\\
    & Ensemble & -183.6 & \textcolor{RawSienna}{\textbf{-213.2}} & -14.9 & -10.1 & -28.6 &-20.6 & -63.2 & -46.2\\
    & Transfer Learning & -132.4 &\textcolor{RawSienna}{\textbf{-89.4}} &-10.7 &-9.1 &-18.5 &-17.3 &-38.7 &-38.2\\
    & TL + Embedding & -172.5 & \textcolor{RawSienna}{\textbf{-138.7}} & -11.6 &-9.5 &-22.7 &-21 &-55.9 &-51.9\\
    & Tuned PPO & -165.6 & \textcolor{RawSienna}{\textbf{-125.5}} & -14.9 & -10.7 & -27.2 & -19.4 & -54.1 & -39.3\\
    & HPPO & -255.6 & \textcolor{RawSienna}{\textbf{-112.7}} & -16.8 &-12.9 &-31.7 &-33.3 &-76.3 &-84.7\Bstrut\\
    \hline
    \multirow{6}{*}{\rotatebox[origin=c]{90}{\shortstack[c]{Scenario 5:\\Path Addition}}} & Multi-Ensemble & -71.2 & \textcolor{RawSienna}{\textbf{-24.5}} & -6 & -6.5 & -9.9 & -10.1 & -19.4 & -19.3\Tstrut\\
    & Ensemble & -74.1 & \textcolor{RawSienna}{\textbf{-26.4}} & -5.7&-6.5 &-11.1 &-10.2 &-20.8 &-19.5\\
    & Transfer Learning & -87.6 & \textcolor{RawSienna}{\textbf{-25.3}} &-8.5 &-6.3 &-14.5 &-10 &-29.8 &-18.3\\
    & TL + Embedding & -103.5 & \textcolor{RawSienna}{\textbf{-43.2}}&-8.8 &-6.3 &-16.5 &-11.6 &-34.7 &-25.5\\
    & Tuned PPO & -74.3 & \textcolor{RawSienna}{\textbf{-1.2}} & -5& -8.1 & -8.5 & -13 & -16.7 & -23.1\\
    & HPPO & -117.4 & \textcolor{ForestGreen}{\textbf{+2.3}} & -4.5 & -8.8 &-7.9 &-22.2 &-16.8 &-57.2\Bstrut\\
    \hline
    \end{tabular}
    \caption{Results from running each of our six submissions to the CAGE Challenge against Scenarios 3, 4, and 5 from Section~\ref{subsec:topology} for 500 episodes of game lengths 30, 50, and 100, including the summed total reward and the percentage change for each agent from their performance in Scenario 2. Percent changes in \textbf{bold} highlight learning agents with the \emph{least} impacted performance. 
    }
    \label{tab:topology}
\end{table*}

In this section we analyze how well our approaches perform when evaluated in an environment different than the one they were trained in.
This comparison helps serve as a proxy for a deployment scenario where the RL agent is trained in a simulator that is mismatched from the actual network it is supposed to defend.
Such a scenario has real-world plausibility: a defender \emph{should} have full knowledge of their environment, and can accurately encode it within a simulator.
On the other hand, in practice proper asset management can be challenging for even sophisticated organizations.

We expect differences between training and evaluation environments to have a significant impact on performance due to the observation structure within CybORG: each bit in the observation refers to a specific quality for a specific host, and thus the defender will learn very \emph{explicit} dynamics for each individual host during training.
Indeed, this problem was highlighted in \cite{applebaum2022bridging} which showed that the bit-vector approach did not generalize to unseen environments when using Tabular Q-Learning.


To test this, we introduce three new scenarios that modify the original \emph{Scenario 2} used in the CAGE Challenge, referring to these as \emph{Scenario 3}, \emph{4}, and \emph{5}.
Each scenario changes the \emph{OS image} -- and corresponding services -- for a select set of hosts and/or changes the \emph{attack path} and the order in which the attacker discovers new hosts.
Scenarios 3 and 4 are both isomorphic with Scenario 2, swapping names between four \emph{User} hosts for Scenario 3, and swapping names between two \emph{Enterprise} servers for Scenario 4.
Both scenarios modify the OS images for these swapped hosts, with Scenario 3 also modifying the next-hop discovery path.
Scenario 5 is not isomorphic to Scenario 2, and instead adds more attack path options for the attacker: namely, in Scenario 2 two of the \emph{User} hosts discover the \emph{Enterprise0} host, and another two \emph{User} hosts discover the \emph{Enterprise1} host.
In Scenario 5, we expand this so that all four \emph{User} hosts discover both of the enterprise hosts.

Table~\ref{tab:topology} provides the results from running each of our approaches against all three scenarios; we tested Heuristic Restore and Sleep and found each of them to score within a few percentage points of their original Scenario 2 score.
Additionally, we do not train any RL-based defenders on these scenarios, as due to the similarities with Scenario 2 we would expect results to be the same.
The results for our approaches are all almost uniformly worse in each of the three new scenarios.
While we anticipated a drop, the magnitude of the difference is significant, particularly for Scenarios 3 and 4 where both ensemble approaches dropped by nearly 200\%.
The only approach that did not have a huge drop in these two scenarios was Tuned PPO, which only fell 29.6\% in Scenario 3.
We credit this to Tuned PPO's Gaussian noise exploration strategy, which in some cases leads to local optima \cite{zhang2022proximal}; here it may be that Tuned PPO fell into such an optima which was more resilient to the observation space mismatch introduced in this scenario.
The only other model that stands out is Transfer Learning, which drops 88.7\% and 89.4\% in Scenarios 3 and 4.
While this is a marked drop in performance, it is the highest scorer for Scenario 4, and is second in score only to Tuned PPO for Scenario 3.

For both Scenarios 3 and 4 the performance against BLine is worse than in the original Scenario 2, likely as a result of BLine taking advantage of the observation mismatch and moving more quickly.
For Scenario 3, the Meander scores are similar to Scenario 2 due to the location of the observation changes -- Meander will compromise all \emph{User} hosts before moving to the \emph{Enterprise} hosts, by the time of which the observation mismatch is not as relevant.
For Scenario 4, however, performance against Meander is notably less than the others, even though this Scenario only modified OS images for two servers and left the attack path unchanged.
This drop is likely due to the learned decoys no longer applying; since decoys are OS specific the swap renders the learned decoy strategy ineffective.


Scenario 5 does not prove as challenging for our approaches, with HPPO and Tuned PPO scoring roughly the same as Scenario 2, and the other four models only performing slightly worse.
This drop appears to come entirely from BLine; the additional attack paths from the \emph{User} hosts to the \emph{Enterprise} ones does not impact Meander as it must compromise all \emph{User} hosts regardless, and so performance is the same as it was in Scenario 2.
By contrast, the BLine attacker will randomly take advantage of the new path, and so we can see performance against it in Scenario 5 slightly worse than Scenario 2.
This is most likely explained by each defensive agent effectively learning optimizations to combat the BLine strategy, with this strategy less effective when BLine is given more options.

We note that Scenarios 3 and 4 are least likely to occur in the real-world as the defender will usually know some facts about each host.
Scenario 5 by contrast is much more likely: each host has the same OS image, but the defender only knows a subset of all attack paths in the network.
Extrapolating towards a future integration, we would expect to see differences between simulation and deployment be more similar to Scenario 5 than Scenarios 3 and 4.

\subsection{Attacker Strategy Robustness}\label{subsec:robust}
\begin{table}\footnotesize
    \centering
    \begin{tabular}{|l|cc|c|c|}
        \hline
        Model & Meander & RandomMeander & Percent Change & Avg. Decision Time (ms)\Tstrut\Bstrut\\
        \hline
        Heuristic Restore & -192.62 & -175.22 & \textcolor{ForestGreen}{\textbf{+10.6}} & 0.006\Tstrut\\
        Random & -566.41 & -428.16 & \textcolor{ForestGreen}{\textbf{+24.4}} & -\\
        Sleep & -972.43 & -794.0 & \textcolor{ForestGreen}{\textbf{+18.3}} & -\Bstrut\\
        \hline
        Multi-Ensemble & -19.45 & -26.83 & \textcolor{RawSienna}{\textbf{-37.9}} & 27.44\Tstrut\\
        Ensemble & -19.81 & -26.34 & \textcolor{RawSienna}{\textbf{-33}} & 6.65\\
        Transfer Learning & -18.24 & -24.68 & \textcolor{RawSienna}{\textbf{-35.3}} & 0.81\\
        TL + Embedding & -24.31 & -26.95 & \textcolor{RawSienna}{\textbf{-10.9}} & 0.71\\
        Tuned PPO & -22.82 & -37.54 & \textcolor{RawSienna}{\textbf{-64.5}} & 0.26\\
        HPPO & -57.89 & -42.08 & \textcolor{ForestGreen}{\textbf{+27.3}} & 0.97\Bstrut\\
        \hline
    \end{tabular}
    \caption{\Tstrut Average total reward and individual decision time for each approach against RandomMeander over 1000 episodes of length 100. Percentage Change shows the relative increase or decrease in reward against RandomMeander as compared to Meander; a positive change indicates better defender performance against RandomMeander, and a negative change indicates worse performance.}
    \label{tab:randommeanderresults}
\end{table}


Our last integration analysis looks at the robustness of each approach against an unseen attacker strategy.
This is inspired in part by \cite{molina2021network} and \cite{hammar2020finding}, which both found that RL-based defenders performed well against static attackers, with degraded performance against an RL-based attacker.
Our new attacker does not use RL, but instead more simply modifies Meander to allow for \emph{duplicate compromise}.
During a game, Meander keeps a list of which hosts it has exploited and is prohibited -- regardless of foothold -- from exploiting any host it has previously exploited, except in cases where the most recent exploit has failed.
This restriction optimizes Meander to avoid exploiting already-compromised hosts, but comes at a cost where it will not re-compromise (select) lost footholds.
We create a new strategy -- \emph{RandomMeander} -- that removes this restriction, allowing the attacker to exploit a random (known) host, regardless if that host has been compromised by the attacker. 



Table~\ref{tab:randommeanderresults} contains the results from running three of the baseline approaches in Table~\ref{tab:baselines} and our new approaches against RandomMeander for 1000 episodes of length 100.
The first three rows of Table~\ref{tab:randommeanderresults} show performance against three of the baselines.
In each case, the baseline defender performs \emph{better} against RandomMeander than it does against Meander, due in part to the former moving more slowly because it wastes turns by exploiting currently compromised hosts.

By contrast, nearly all of our models perform \emph{worse} against RandomMeander than they do against Meander.
Multi-Ensemble, Ensemble, and Transfer Learning all drop over 30\%, with TL + Embedding also performing worse, but only 10.9\%.
These are each large drops, but they only amount to a small number of points (\textasciitilde 7) and their scores still remain convincingly higher than the baselines.
Interestingly, despite these models varying in performance against Meander, their scores against RandomMeander are within \textasciitilde 2 points of one another.
Tuned PPO has an even bigger performance drop, scoring over 60\% worse against RandomMeander as opposed to Meander.
HPPO differs in that it performs better against RandomMeander: its tiered approach may offer more robustness against attacker variation as it is not as optimized against each attacker strategy's nuances.
We do note that the timing against RandomMeander is similar to the timing for BLine and Meander as per Table~\ref{tab:timing}.



Part of RandomMeander's relative success against our defenders is likely explained by it following different paths than Meander: RandomMeander will try to \emph{re-compromise} lost footholds, a strategy our agents have not seen.
Looking at trends across agents, we also see that our agents perform better against more deterministic agents -- against the baselines, BLine, then Meander, then RandomMeander are best, but against our approaches, RandomMeander, Meander, then BLine are best.
We note that even though the reward difference between Meander and RandomMeander is small, it may imply the existence of an untested attacker strategy that is even more effective against our defenders.
\section{Discussion and Future Work}\label{sec:future}
We believe that ensemble models could be quite powerful for future autonomous network defense solutions, with our ensembles outperforming other approaches in this work likely due to learning a better decoy strategy (Appendix~\ref{app:actdist}).
For the future, alternative ensemble voting techniques -- e.g., Boltzmann multiplication \cite{wiering2008ensemble} or other heuristics based on constituent model properties -- could provide better robustness or performance. 
Another approach is to include stronger individual models in the ensemble, or using non-PPO models within an ensemble to offer better model diversity and robustness.
Additionally, leveraging the results of \cite{foley2022autonomous} and using an ensemble of HiPPO model or building a HiPPO model that uses ensembles as sub-policies could be promising.

The larger future challenges stem from safely integrating autonomous RL network defense agents with complex live data and high-consequence security playbooks.
While there are numerous sub-challenges from both the environment and integration, we highlight: environment generalization, attacker-defender robustness, human-machine teaming, and insightful measures of system effectiveness. 

Generalization to unseen environments is sparsely tested in the literature, with many if not all of the approaches previously mentioned \cite{zonouz2013rre, musman2019steps, molina2021network, foley2022autonomous, hammar2020finding} likely to suffer performance losses similar to those tested here.
Recent work in \cite{applebaum2022bridging} modifies CyberBattleSim \cite{team2021cyberbattlesim} to add a defender interface and allow for reasoning about host features instead of individual hosts, testing different state space designs.
They find that the bit-vector approach used in this work struggles to generalize to unseen environments, but approaches that use more abstract state spaces -- e.g., binning the percentage of compromised hosts -- offer much better performance in unseen environments.
Another recent approach \cite{collyeracd} suggests modifying the observation space to use graph-based features to augment the traditional bit-vector representation and support better generalization.


Robustness against attacker strategies also remains an open problem, noted in other autonomous network defense research \cite{molina2021network, hammar2020finding}, the latter of which uses \emph{self-play} \cite{silver2016mastering} and \emph{opponent pools} \cite{berner2019dota} but still struggles to converge on an optimal policy against a dynamic attacker.
The authors of \cite{hammar2020finding} show in \cite{hammar2022learning} that by re-framing the network defense problem as one of optimal stopping \cite{dynkin1969game} they can use a game-theoretic solution that is more robust against dynamic attackers.
Regardless, future approaches using RL will need to consider robustness against attacker strategy given (1) the adversarial nature of security domain -- which puts all system components, including the RL agent itself, in scope -- and (2) the potential susceptibility of RL deployments to adversarial examples and other attacks \cite{lin2017tactics}.


Lastly, we note that any deployed RL-based defensive agent should be measured against how effective it is in relation to \emph{human operators}.
Prior work has looked at proving optimal solutions \cite{musman2019steps, hammar2022learning} or automated-only comparative analysis \cite{molina2021network}, though \cite{prebotcognitive} recently compared automated defender and human performances, and \cite{holm2022lore} ran an end-user study showing usability of an automated attacker.


 \section{Conclusion}\label{sec:discussion}
In this work we examined several RL techniques for autonomous network defense within a high-fidelity network security simulator, as defined in the second edition of the CAGE Challenge.
Of our approaches, leveraging an ensemble of PPO-based policies and using a majority voting scheme was most effective.
This strategy outperformed our other approaches, ranked second place in the CAGE competition, and scored similarly to an RL-based defender with full network visibility.
We believe ensemble approaches will push the state of the art for autonomous network defense agents. 

To understand the practical viability of RL-based approaches, we also explored challenges to real-world integration for autonomous network defense, including timing, generalization to unseen environments, and robustness to attacker strategy.
Timing did not appear to be a factor for this set of methods, with all models performing within reasonable bounds. 
Though the unseen environments differed only slightly from the training environment, all of our approaches performed significantly worse, with the ensemble approaches degraded the most; we observed that modifying hosts caused the most disruption to our defenders, with adding attack paths not being as impactful.
For robustness against alternative attacker strategies, we discovered that testing against an unseen but \emph{less} efficient attacker in fact was \emph{harder} for our approaches to defend against, likely due to the unseen attacker being more unpredictable and executing attack paths previously unseen.
Combining these conclusions, though RL approaches can create effective defenders in some limited scenarios, integration to a production system will require significant future work and investment. 

\bibliographystyle{plainnat}
\bibliography{bib.bib}

\appendix
\section{Additional Details}\label{app:params}
This section describes each approach's specific parameters and training setup; missing parameters should be assumed to align with the RLlib or Stable Baselines3 default.
We also include a visualization and discussion of the spread for each model's results from Section~\ref{sec:results} -- including comments about seed sensitivity -- as well as analysis of each model's action distribution.

\subsection{Variance and Seed Sensitivity}\label{app:variance}
\begin{figure*}[h!]
    \centering
    \includegraphics[width=.99\textwidth]{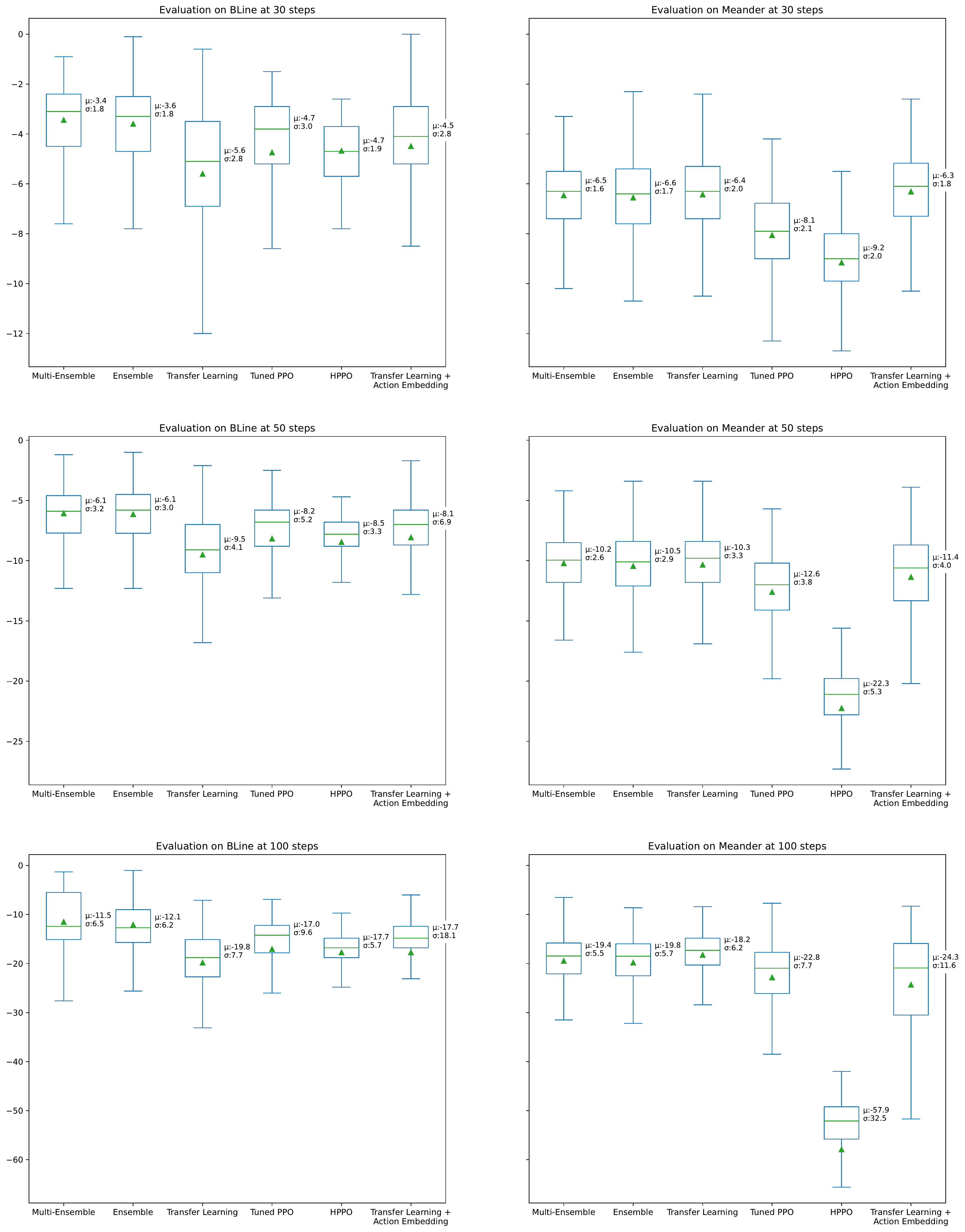}
    \caption{Boxplots showing spread for rewards from select models tested in Table~\ref{tab:finalresults}. The box spans the range of the 25th percentile to 75th percentile, with the green line showing the median. Additionally, the green triangle shows the average, with the side annotation of $\mu$ showing the mean and $\sigma$ the standard deviation.}
    \label{fig:test}
\end{figure*}

Figure~\ref{fig:test} shows the average total rewards from Table~\ref{tab:finalresults} alongside their standard deviations as well as the 25th to 75th percentiles.
This visualization highlights the high degree of overlap in each models' results -- aside from the results for HPPO against Meander, many of the models' average total reward was within range of the others.
We found this overlap to remain even when increasing the number of trials: running several of the models at 5000 trails as opposed to 1000 resulted in the same variance.

We ultimately credit the spread to \emph{seed sensitivty} within CybORG itself: the seed impacts the specific attack mechanics that the red agent uses, which in turn impacts each decoy's efficacy.
By contrast, the organizers of the CAGE Challenge fix the seed to a constant value during evaluation, with each of our approaches having a much smaller variance; i.e. Multi-Ensemble had the smallest with a 95\% confidence interval of $\pm 0.59$ and HPPO had the largest at $\pm 1.63$.
For future work we plan to better analyze the variance and the approach's sensitivity to the seed value, although anecdotally we found that the relative performance of each model stayed consistent across seeds.

\subsection{Action Distribution}\label{app:actdist}
\begin{table}[h!]\footnotesize
    \centering
    \begin{tabular}{|l|c|c|c|c|}
    \hline
    Approach & Wrong Restores & Wrong Removes & Bad Host Targeting & Sleep\Tstrut\Bstrut\\
    \hline
    Multi-Ensemble & 0.23 & 1.4 & 4.72  & 0.7\Tstrut\\
    Ensemble & 0.49 & 2.46 & 4.16 & 1.18 \\
    Transfer Learning & 0.28 & 6.27 & 3.66 & 0\\
    TL + Embedding & 0.33 & 6 & 0.91 & 0 \\
    Tuned PPO & 0.6 & 1.81 & 11.04 & 1.19 \\
    HPPO & 0.44 & 0 & 0 & 0\Bstrut\\
    \hline
    \end{tabular}
    \caption{\Tstrut Average number of incorrect actions per game by methodology, averaged over 500 games of length 100 against each of BLine and Meander.} 
    \label{tab:unhelpfulactions}
\end{table}

We sought to analyze the number of ``correct'' actions each model chose during a game.
Lacking an optimal policy to compare to, we instead identified four actions and context where it is clear the action is clearly \emph{wrong} or at least \emph{unhelpful}:
\begin{itemize}
    \item \emph{Wrong Restores.} The defender restores a host that is not compromised, incurring a penalty.
    \item \emph{Wrong Removes.} The defender attempts to remove malware from a non-compromised host.
    \item \emph{Bad Host Targeting.} The defender acts on a host the attacker is \emph{unable} to scan/interact with.
    \item \emph{Sleep.} The defender executes a no-op.
\end{itemize}

Table~\ref{tab:unhelpfulactions} shows the average number of each category for each of our approaches, averaged over 500 games of length 100 against each of BLine and Meander.
Despite each approach having highly varied overall performances from each other, we see that each averages roughly the same number of incorrect Restore actions.
We suspect this is due to the penalty for the Restore action, where each model learns more explicitly to avoid incorrect applications of this action.

The other three categories have varied spreads for each approach.
Both Ensemble approaches have similar average number of incorrect Remove and Sleep actions, the former of which is to be expected due to noise and the latter from particular evaluation settings.
Both Ensembles average \textasciitilde 4 actions that target the unreachable hosts, which we were surprised to see given the attacker always ignores these.
Tuned PPO has a similar profile, with a low number for incorrect Remove and Sleep actions, but in this case a much higher number of actions targeting the unreachable hosts.
The two Transfer Learning approaches notably have a higher average number of incorrect Remove actions, but do not use Sleep and have fewer actions targeting the unreachable hosts.

\begin{sidewaysfigure*}[h!]
    \centering
    \includegraphics[width=.99\textwidth]{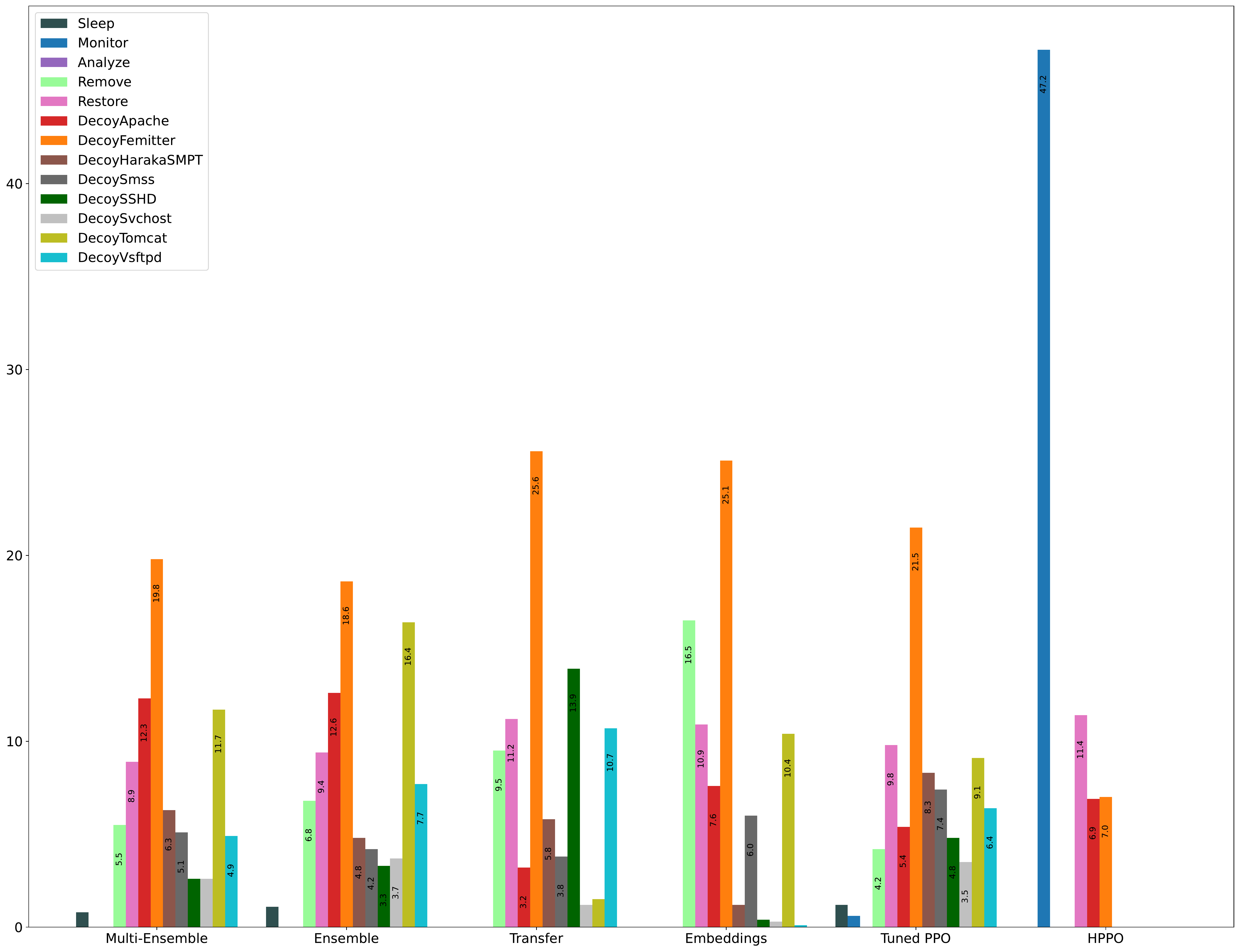}
    \caption{Bar chart showing the average number of actions per action type for each of the six models submitted to the CAGE Challenge. Numbers averaged over 500 episodes, combining results for both BLine and Meander at game length of 100.}
    \label{fig:action_dist}
\end{sidewaysfigure*}

Ultimately we did not feel this analysis proved helpful as a proxy for overall performance.
Looking at HPPO we see no incorrect Removes, targeting unreachable hosts, or Sleep, but it has the worst overall score among our approaches.
Instead, the success of each approach seems more tightly coupled with the specific action distribution used by each approach, visualized in Figure~\ref{fig:action_dist}.
Looking at the figure shows that HPPO overprioritizes the \emph{Monitor} action, while the other five approaches all favor using decoys.
The nuance and difference however between these approaches seems to fall more on the side of decoy deployment strategy, with each approach preferring different decoys to use.
We suspect that the particular placement of these decoys is an important factor as well, and believe that future work trying to operationalize these ideas should investigate better explaining particular model decisions.

\begin{table}[b!]\footnotesize
    \centering
    \begin{tabular}{|c|c|c|c|c|c|}
        \hline
        Model & Iteration & $Pr(\texttt{BLine})$ & $Pr(\texttt{Meander})$& Game Length & Steps\Tstrut\Bstrut\\
        \hline
        Transfer Learning & 1 & 0 & 1 & 30 & 400000\Tstrut\\
        & 2 & 1 & 0 & 30 & 400000\\
        & 3 & 0 & 1 & 50 & 400000\\
        & 4 & 1 & 0 & 50 & 400000\\
        & 5 & 0 & 1 & 100 & 800000\\
        & 6 & 1 & 0 & 100 & 800000\\
        & 7 & 0 & 1 & 100 & 400000\Bstrut\\
        \hline
        TL + Embedding & 1 & 0.95 & 0.05 & 30 & 500000\Tstrut\\
        & 2 & 0.05 & 0.95 & 30 & 500000\\
        & 3 & 0.95 & 0.05 & 50 & 500000\\
        & 4 & 0.05 & 0.95 & 50 & 500000\\
        & 5 & 0.95 & 0.05 & 100 & 800000\\
        & 6 & 0.05 & 0.95 & 100 & 800000\\
        & 7 & 0 & 1 & 100 & 400000\Bstrut\\
        \hline
    \end{tabular}
    \caption{\Tstrut Training schedules for the Transfer Learning and TL + Embedding.}
    \label{tab:trainingschedules}
\end{table}


\begin{sidewaystable*}\footnotesize
    \centering
    \begin{tabular}{|c||l|c||l|c|}
        \hline
        Model  & Parameter & Value & Parameter & Value\Tstrut \Bstrut\\
        \hline
        \multirow{6}{*}{\shortstack[c]{Tuned PPO}} & \texttt{activation\_fn} & \texttt{relu} & \texttt{batch\_size} & 16\Tstrut\\
        & \texttt{clip\_range} & 0.1 & \texttt{entropy\_coef} & 0.0002\\
        & \texttt{gae\_lambda} & 0.99 & \texttt{gamma} & 0.99\\
        & \texttt{learning\_rate} & $5.0148 \cdot 10^{-5}$ & \texttt{max\_grad\_norm} & 0.5\\
        & \texttt{number\_epochs} & 10 & \texttt{number\_steps} & 2048\\
        & \texttt{vf\_coeff} & 0.102 & &\Bstrut\\
        \hline
        Full Visibility: Meander & \texttt{entropy\_coeff} & 0.0001 & \texttt{vf\_clip\_param} & 100\Tstrut\Bstrut\\
        \hline
        Full Visibility: BLine & \texttt{gamma} & 0.95 & \texttt{entropy\_coeff} & 0.0001\Tstrut\Bstrut\\
        \hline
        \hline
        \multicolumn{1}{|c||}{}& \multicolumn{1}{c|}{Parameter} & \multicolumn{3}{c|}{Value}\Tstrut\Bstrut\\\cline{1-5}
        \multirow{7}{*}{\shortstack[c]{Transfer Learning}} &  \texttt{entropy\_coeff\_schedule} & \multicolumn{3}{c|}{[[0, 0.001], [1000.0, 0.0001], [10000.0, $10^{-5}$], [100000.0, $10^{-6}$]]}\Tstrut\\
        & \texttt{exploration\_config.feature\_dim} & \multicolumn{3}{c|}{512}\\
        & \texttt{exploration\_config.forward\_net\_hiddens} & \multicolumn{3}{c|}{[512]}\\
        & \texttt{exploration\_config.lr} & \multicolumn{3}{c|}{$10^{-4}$}\\
        & \texttt{lr} & \multicolumn{3}{c|}{$9.10^{-5}$}\\
        & \texttt{kl\_coeff} & \multicolumn{3}{c|}{0.3}\\
        & \texttt{model.fcnet\_hiddens} & \multicolumn{3}{c|}{[128, 128]}\Bstrut\\
        \hline
        \multirow{7}{*}{\shortstack[c]{TL + Embedding}} & \texttt{entropy\_coeff\_schedule} & \multicolumn{3}{c|}{[[0, 0.001], [1000.0, 0.0001], [10000.0, $10^{-5}$], [100000.0, $10^{-6}$]]}\Tstrut\\
        & \texttt{exploration\_config.feature\_dim} & \multicolumn{3}{c|}{512}\\
        & \texttt{exploration\_config.forward\_net\_hiddens} & \multicolumn{3}{c|}{[512]}\\
        & \texttt{exploration\_config.lr} & \multicolumn{3}{c|}{$10^{-4}$}\\
        & \texttt{lr} & \multicolumn{3}{c|}{$9.10^{-5}$}\\
        & \texttt{kl\_coeff} & \multicolumn{3}{c|}{0.3}\\
        & \texttt{model.fcnet\_hiddens} & \multicolumn{3}{c|}{[128, 128]}\Bstrut\\
        \hline
        \multirow{6}{*}{\shortstack[c]{HPPO}} & \texttt{entropy\_coeff\_schedule} & \multicolumn{3}{c|}{[[0, 0.001], [1000.0, 0.0001], [10000.0, $10^{-5}$], [100000.0, $10^{-6}$]]}\Tstrut\\
        & \texttt{exploration\_config.feature\_dim} & \multicolumn{3}{c|}{128}\\
        & \texttt{exploration\_config.lr} & \multicolumn{3}{c|}{0.001}\\
        & \texttt{lr} & \multicolumn{3}{c|}{$9.10^{-6}$}\\
        & \texttt{model.policies.high\_level\_policy.gamma} & \multicolumn{3}{c|}{0.95}\\
        & \texttt{model.policies.low\_level\_policy.gamma} & \multicolumn{3}{c|}{0.99}\Bstrut\\
        \hline
    \end{tabular}
    \caption{\Tstrut Hyper-parameter values for various models.}
    \label{tab:allhyperparameters}
\end{sidewaystable*}

\begin{sidewaystable*}\footnotesize
    \centering
    \begin{tabular}{|l|lc|l|c|ccccc|}
    \hline
        \multirow{2}{*}{\shortstack[c]{Ensemble\\ID + Score}} & \multirow{2}{*}{Model} & \multirow{2}{*}{Score} & \multirow{2}{*}{Strategy} & \multirow{2}{*}{\shortstack[c]{Training\\Steps}} & \multirow{2}{*}{$Pr(\texttt{BLine})$} & \multirow{2}{*}{$Pr(\texttt{Meander})$} & \multirow{2}{*}{\texttt{gamma}} & \multirow{2}{*}{\texttt{entropy\_coeff}} & \multirow{2}{*}{\texttt{model.fcnet\_hiddens}}\Tstrut\Bstrut\\
        & & & & & & & & &\\
        \hline
        1: -59.37 & 0567a5ffcd1e458fba7bdfa385f299c3 & -79.98 & Normal & 3.0mm & 0.25 & 0.75 & 0.95 & 0.0001 & [256, 256]\Tstrut\\
        & 3aaa5ff3de5a4b19bac0861e83982e91 & -72.66 & Normal & 1.6mm & 0.5 & 0.5 & 0.96 & 0 & [256, 256]\\
        & 541b39af6d0d477cb6b535fa9356e3a0 & -77.26 & Maintenance & 1.6mm & 0.5 & 0.5 & 0.95 & 0 & [256, 256]\\
        & bc1b3cdd5a704f01be7c6f8847cded57 & -78.96 & Normal & 3.0mm & 0.5 & 0.5 & 0.925 & 0 & [256, 256]\\
        & c1dc05deffde4d3b9ab9af338844f756 & -73.06 & Normal & 3.0mm & 0.5 & 0.5 & 0.95 & 0 & [256, 256]\\
        & d03ff193281e4043862c0232779d3e58 & -81.46 & Maintenance & 1.6mm & 0.5 & 0.5 & 0.95 & 0.0001 & [256, 256]\\
        & ddc46c43b85e4ffc87e18ac6986c1850 & -77.47 & Normal & 3.0mm & 0.75 & 0.25 & 0.95 & 0.001 & [256, 256]\Bstrut\\
        \hline        
        2: -58.69 & 3aaa5ff3de5a4b19bac0861e83982e91 & -72.66 & Normal & 1.6mm & 0.5 & 0.5 & 0.96 & 0 & [256, 256]\Tstrut\\
        & 541b39af6d0d477cb6b535fa9356e3a0 & -77.26 & Maintenance & 1.6mm & 0.5 & 0.5 & 0.95 & 0 & [256, 256]\\
        & b296b7eb79a240e3bf47fd4742146682 & -74.29 & Normal & 3.0mm & 0.75 & 0.25 & 0.95 & 0.0001 & [256, 256]\\
        & bc1b3cdd5a704f01be7c6f8847cded57 & -78.96 & Normal & 3.0mm & 0.5 & 0.5 & 0.925 & 0 & [256, 256]\\
        & c1dc05deffde4d3b9ab9af338844f756 & -73.06 & Normal & 3.0mm & 0.5 & 0.5 & 0.95 & 0 & [256, 256]\\
        & d03ff193281e4043862c0232779d3e58 & -81.46 & Maintenance & 1.6mm & 0.5 & 0.5 & 0.95 & 0.0001 & [256, 256]\\
        & ed7011ae66ae4f5fac28b384f0c4cb00 & -76.57 & Normal & 1.6mm & 0.5 & 0.5 & 0.95 & 0.0001 & [256, 256]\Bstrut\\
        \hline
        3: -58.76 & 0567a5ffcd1e458fba7bdfa385f299c3 & -79.98 & Normal & 3.0mm & 0.25 & 0.75 & 0.95 & 0.0001 & [256, 256]\Tstrut\\
        & 3aaa5ff3de5a4b19bac0861e83982e91 & -72.66 & Normal & 1.6mm & 0.5 & 0.5 & 0.96 & 0 & [256, 256]\\
        & 541b39af6d0d477cb6b535fa9356e3a0 & -77.26 & Maintenance & 1.6mm & 0.5 & 0.5 & 0.95 & 0 & [256, 256]\\
        & c1dc05deffde4d3b9ab9af338844f756 & -73.06 & Normal & 3.0mm & 0.5 & 0.5 & 0.95 & 0 & [256, 256]\\
        & bc1b3cdd5a704f01be7c6f8847cded57 & -78.96 & Normal & 3.0mm & 0.5 & 0.5 & 0.925 & 0 & [256, 256]\\
        & d03ff193281e4043862c0232779d3e58 & -81.46 & Maintenance & 1.6mm & 0.5 & 0.5 & 0.95 & 0.0001 & [256, 256]\\
        & ddc46c43b85e4ffc87e18ac6986c1850 & -77.47 & Normal & 3.0mm & 0.75 & 0.25 & 0.95 & 0.001 & [256, 256]\Bstrut\\
        \hline
        4: -60.15 & 3aaa5ff3de5a4b19bac0861e83982e91 & -72.66 & Normal & 1.6mm & 0.5 & 0.5 & 0.96 & 0 & [256, 256]\Tstrut\\
        & 5116aaeeb8234c0980fa0ef4fa409cf0 & -117.53 & Normal & 3.0mm & 0.95 & 0.05 & 0.9 & 0 & [256, 256]\\
        & 541b39af6d0d477cb6b535fa9356e3a0 & -77.26 & Maintenance & 1.6mm & 0.5 & 0.5 & 0.95 & 0 & [256, 256]\\
        & 62f271a5b26a41e2801bd0ed5316f98b & -474.3 & Normal & 1.5mm & 1 & 0 & 0.95 & 0 & [128, 128]\\
        & b296b7eb79a240e3bf47fd4742146682 & -74.29 & Normal & 3.0mm & 0.75 & 0.25 & 0.95 & 0.0001 & [256, 256]\\
        & c1dc05deffde4d3b9ab9af338844f756 & -73.06 & Normal & 3.0mm & 0.5 & 0.5 & 0.95 & 0 & [256, 256]\\
        & cdf3076e074c4c33b6c538496bd46ed7 & -125.68 & Normal & 3.0mm & 0.95 & 0.05 & 0.95 & 0 & [256, 256]\\
        & ed7011ae66ae4f5fac28b384f0c4cb00 & -76.57 & Normal & 1.6mm & 0.5 & 0.5 & 0.95 & 0.0001 & [256, 256]\Bstrut\\
        \hline
        5: -59.37 & 0567a5ffcd1e458fba7bdfa385f299c3 & -79.98 & Normal & 3.0mm & 0.25 & 0.75 & 0.95 & 0.0001 & [256, 256]\Tstrut\\
        & 157bde6d96e24388b6d1fe6e6e487e85 & -109.25 & Normal & 3.0mm & 0.75 & 0.25 & 0.9 & 0.0001 & [256, 256]\\
        & 3aaa5ff3de5a4b19bac0861e83982e91 & -72.66 & Normal & 1.6mm & 0.5 & 0.5 & 0.96 & 0 & [256, 256]\\
        & 5116aaeeb8234c0980fa0ef4fa409cf0 & -117.53 & Normal & 3.0mm & 0.95 & 0.05 & 0.9 & 0 & [256, 256]\\
        & b5c4ca39a6ba4ebab3b1c4c7939427bd & -78.66 & Normal & 3.0mm & 0.75 & 0.25 & 0.95 & 0 & [256, 256]\\
        & c1dc05deffde4d3b9ab9af338844f756 & -73.06 & Normal & 3.0mm & 0.5 & 0.5 & 0.95 & 0 & [256, 256]\\
        & ddc46c43b85e4ffc87e18ac6986c1850 & -77.47 & Normal & 3.0mm & 0.75 & 0.25 & 0.95 & 0.001 & [256, 256]\\
        & ed7011ae66ae4f5fac28b384f0c4cb00 & -76.57 & Normal & 1.6mm & 0.5 & 0.5 & 0.95 & 0.0001 & [256, 256]\Bstrut\\
        \hline
    \end{tabular}
    \caption{\Tstrut Details for each of the five ensembles. Each ensemble is listed along with its score -- used by the Multi-Ensemble method -- and its constituent models. In addition to the hyper-parameter listed in the table, all monitors also use the following: \texttt{framework} = \texttt{torch}, \texttt{vf\_clip\_param} = 100, and \texttt{train\_batch\_size} = 1000. Note that all models were trained with RLlib. The \emph{Maintenance} strategy refers to a strategy where the model was trained specifically to keep the intermediate reward greater than -10.}
    \label{tab:allensembles}
\end{sidewaystable*}

\end{document}